\documentclass[10pt,journal,compsoc]{IEEEtran}
\usepackage{times}
\usepackage{epsfig}
\usepackage{graphicx}

\usepackage{amssymb}
\usepackage{multirow}
\usepackage{multicol}
\usepackage{subfigure}

\usepackage{algorithm}
\usepackage{algorithmic}

\usepackage{titlesec}
\usepackage{bbding}

\usepackage{amsmath,amssymb,amsfonts}
\usepackage[pagebackref=true,breaklinks=true,colorlinks,bookmarks=false]{hyperref}

\ifCLASSOPTIONcompsoc
  \usepackage[nocompress]{cite}
\else
  \usepackage{cite}
\fi
\ifCLASSINFOpdf
\else
\fi
\begin{document}
%
\title{HAMIL: Hierarchical Aggregation-Based Multi-Instance Learning \\ for Microscopy Image Classification}
%
%
%
%

\author{Yanlun~Tu,
        Houchao~Lei,
        Wei~Long,
        and~Yang~Yang

\IEEEcompsocitemizethanks{
\IEEEcompsocthanksitem 
Y. Tu, H. Lei, W. Long, Y. Yang are with the Department of Computer Science and Engineering,
Shanghai Jiao Tong University, Shanghai, SH 200240, China. E-mail: \{tuyanlun, leihouchao\}@sjtu.edu.cn, yilang2007lw@gmail.com, yangyang@cs.sjtu.edu.cn.
}

\thanks{Manuscript received April 19, 2005; revised August 26, 2015.\\
(Corresponding author: Y. Yang).

}}

%
%

\markboth{Journal of \LaTeX\ Class Files,~Vol.~14, No.~8, August~2015}%
{Shell \MakeLowercase{\textit{et al.}}: Bare Demo of IEEEtran.cls for Computer Society Journals}
%



\IEEEtitleabstractindextext{%
\begin{abstract}
Multi-instance learning is common for computer vision tasks, especially in biomedical image processing. Traditional methods for multi-instance learning focus on designing feature aggregation methods and multi-instance classifiers, where the aggregation operation is performed either in feature extraction or learning phase. As deep neural networks (DNNs) achieve great success in image processing via automatic feature learning, certain feature aggregation mechanisms need to be incorporated into common DNN architecture for multi-instance learning. Moreover, flexibility and reliability are crucial considerations to deal with varying quality and number of instances. 

In this study, we propose a hierarchical aggregation network for multi-instance learning, called HAMIL. The hierarchical aggregation protocol enables feature fusion in a defined order, and the simple convolutional aggregation units lead to an efficient and flexible architecture. We assess the model performance on two microscopy image classification tasks, namely protein subcellular localization using immunofluorescence images and gene annotation using spatial gene expression images. The experimental results show that HAMIL outperforms the state-of-the-art feature aggregation methods and the existing models for addressing these two tasks. The visualization analyses also demonstrate the ability of HAMIL to focus on high-quality instances.
\end{abstract}

}

\maketitle

\IEEEdisplaynontitleabstractindextext

%
\IEEEpeerreviewmaketitle

\IEEEraisesectionheading{\section{Introduction}\label{sec:introduction}}

%
%
%
%
\IEEEPARstart{A} lot of computer vision tasks exhibit multi-instance property, i.e. each input data sample is represented by a bag of instances and learning is performed at the bag level instead of the instance level \cite{zhou2004multi,foulds2010review}. For example, video-based identity recognition can be treated as a multi-instance learning (MIL) task, as each video input is a time-series of frames and each frame can be regarded as an instance \cite{ref24}. Another example is the automatic diagnosis system based on magnetic resonance imaging (MRI) data, which consists of multiple slices presenting the area of interest being scanned \cite{LIU2018157}. In these two cases, there is either temporal or spatial dependency between images; whereas in a lot more MIL scenarios, instances within each bag are unordered or exhibit no dependency among each other. Such unordered MIL tasks are especially common in biomedical image processing \cite{li2012drosophila, YAO2020101789, ilse2018attention,10.1093/bioinformatics/btw252}. 

\begin{figure}[t]
\begin{center}
\includegraphics[width=1\linewidth]{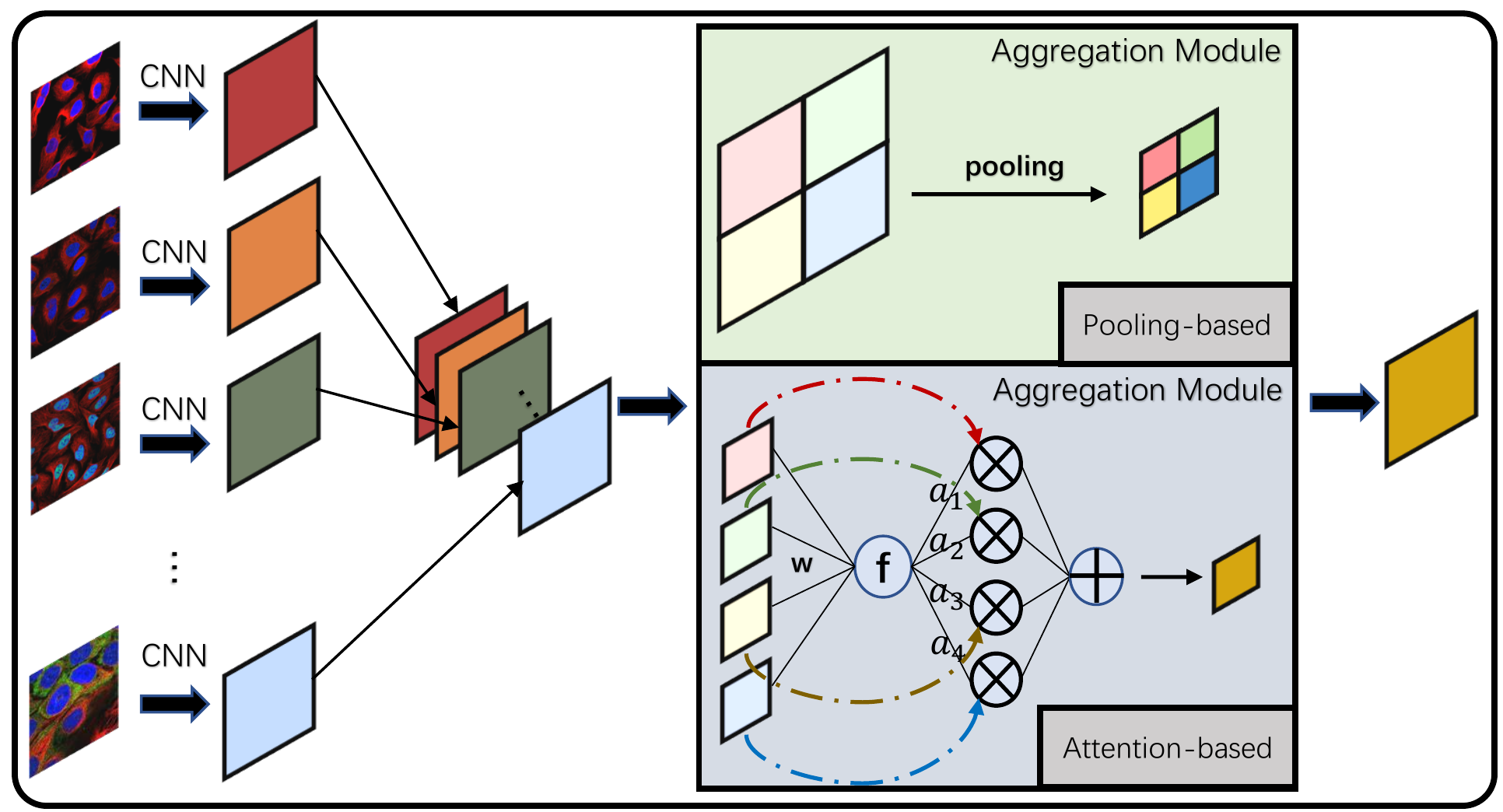}
\end{center}
\caption{Two typical aggregation mechanisms in deep learning models. CNNs are commonly used to extract features from input images, and then the extracted features are fed into the feature aggregation module to obtain the final aggregation output. }
\label{fig:onecol}
\end{figure}

The imaging of biological samples, plays a key role in current life science research, enabling scientists to analyze from tissue level down to the cell level. 
Unlike natural images, biological microscopy images are often much harder to obtain due to the difficulty in preparing the specimen or stringent experimental conditions. Therefore, during the imaging procedure, it is common to capture multiple images for a specimen in a single trial of experiments and perform multiple trials for repeatability. To infer the functions of genes or characteristics of molecules, all the captured images should be considered comprehensively to give a more accurate judgement on the final output, as single images may only contain partial information.

Compared with video or 3D data, dealing with unordered input requires the model to have the permutation-invariant property \cite{ilse2018attention}, i.e., the model should be insensitive to the presenting order of input instances. Moreover, the models also need to address the issues coming from variable-sized inputs and different image qualities. Thus, developing MIL models is a very challenging job. 

Traditional MIL methods fall into two categories, focusing on feature extraction and classification, respectively. 


The first category of methods aggregates multiple instances into a fixed-length input before feeding the classifiers. The aggregation operation can be performed before or after feature extraction. FlyIT \cite{ref27} is an example of the former type, which first stitches images belonging to the same bag into large images and then extracts features from the large images. 
The latter type first extracts features for raw instances and then performs the aggregation operation (as illustrated in Fig. \ref{fig:onecol}). For instance, MIMT-CNN \cite{MIMT-CNN} employs 10 VGG models to learn feature representation from 10 images, respectively, and concatenates them into a bag representation for the following classification. 
The feature aggregation can be regarded as a part of feature engineering, and the subsequent learning procedure remains unchanged. 
Alternatively, common classifiers can be modified to handle inputs of multiple instances. Till now, a lot of MIL methods have been developed, like multi-instance KNN \cite{wang2000solving}, multi-instance neural networks \cite{zhang2004improve,zhou2002neural}, multi-instance decision tree \cite{chevaleyre2001solving}, and multi-instance support vector machines (SVMs) \cite{andrews2003support}. The core idea is to compute pair-wise similarity or loss function at the bag level instead of instance level and define the loss function according to the prediction accuracy on bags. 

Note that before using these traditional learning models, image features should be extracted separately, while deep neural networks perform feature learning automatically. For the past decade, MIL tasks have also benefitted a lot from deep learning \cite{ref23}. 
However, the performance of the existing multi-instance DNN models has been limited due to the high complexity of input data. For one thing, some methods set a fix-length of input \cite{MIMT-CNN, ref27}, which lacks flexibility of handing varying-sized input. For the other thing, since the number of instances per sample is often large while high-quality instances are few, the methods that are unable to identify informative instances may not work well. 

To address these challenges, we propose a model, called HAMIL (\underline{H}ierarchically \underline{A}ggregation for \underline{M}ulti-\underline{I}nstance \underline{L}earning). The model is featured by a hierarchical aggregation protocol with simple but effective aggregation units. It can not only learn input bags with varying size but also give preference to informative instances. We assess the performance of HAMIL on two microscopy image classification tasks. On both tasks, HAMIL achieves significant improvement in prediction accuracy compared with other feature aggregation methods and state-of-the-art predictors for these two tasks.

\section{Related work}
\subsection{Traditional feature aggregation}
In traditional image processing, feature aggregation aims to fuse features extracted from multiple images into a comprehensive feature representation before feeding into a learning model. There are three typical feature fusion methods, the bag of visual words (BOV) \cite{bov}, Fisher vector \cite{fishervector}, and vector of locally aggregated descriptors (VLAD) \cite{vlad}. BOV regards image features as words, builds a vocabulary of local image features and generates a vector of their occurrence counts. The Fisher vector method stores the mixing coefficients of the Gaussian mixture model (GMM) as well as the mean and covariance deviation vectors of the individual components. The VLAD method computes the distance of each feature point to the cluster center closest to it. All of these three methods produce a fixed-length feature vector for the input image set, which can work with traditional machine learning models, like SVMs.

\begin{figure}[t]
\begin{center}
\includegraphics[width=0.9\linewidth]{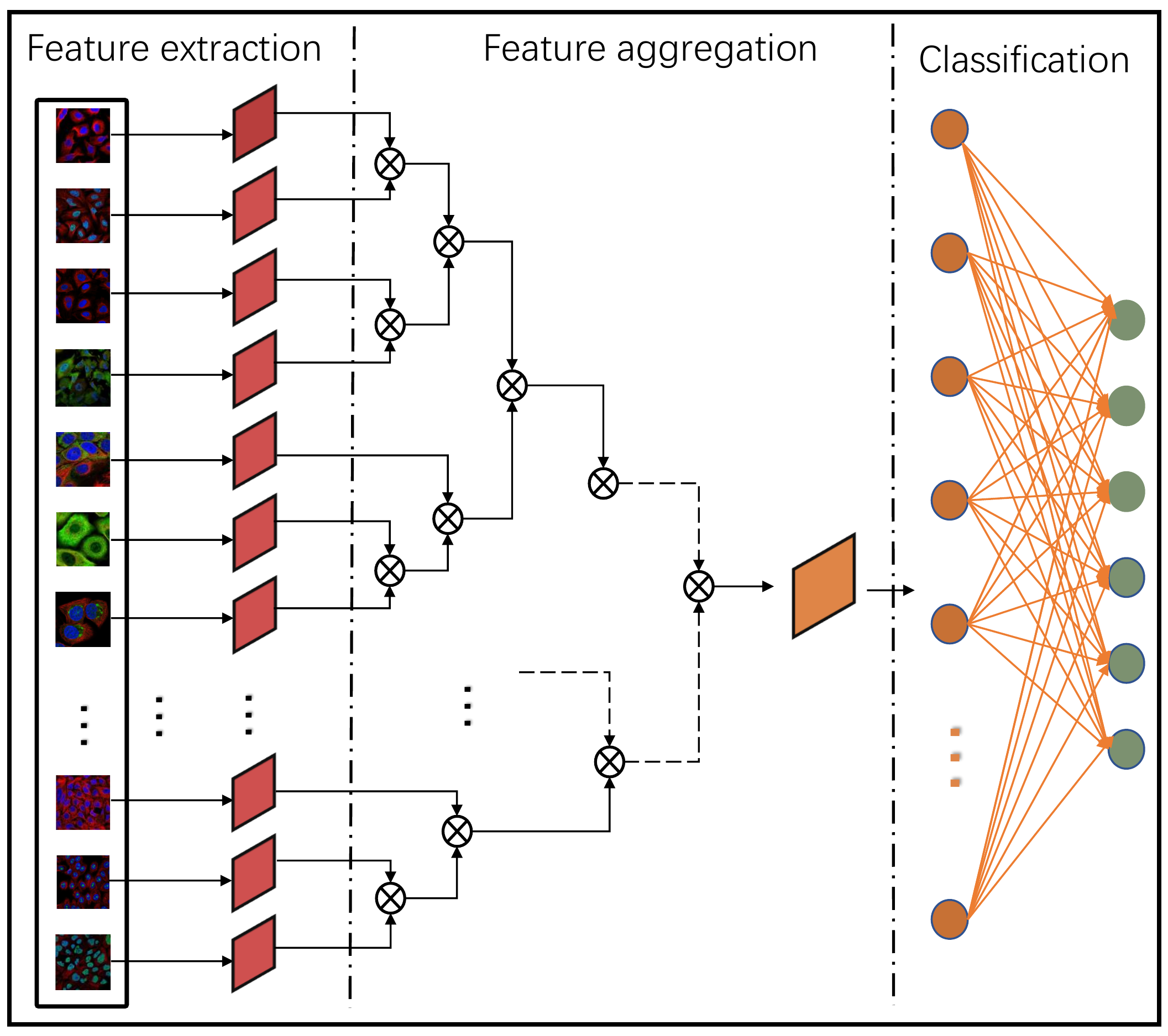}
\end{center}
   \caption{Model architecture of HAMIL. The raw images are first fed to convolutional layers to get feature embeddings and then aggregated based on the hierarchical tree to get a unified encoding for final classification.}
\label{fig:framework}
\end{figure}

\subsection{Aggregation of deep convolutional features}
For the past decade, feature encoding via deep neural networks has almost replaced traditional feature engineering in image processing \cite{ref16} \cite{ref19} \cite{ref20}. Various feature aggregation methods working with deep neural networks have also emerged. There are two typical aggregation schemes, i.e. pooling-based and attention-based. 

As a straightforward aggregation method, the pooling function is permutation-invariant and requires no (or very few) parameters, thus has been widely used in MIL tasks. Babenko \textit{et al.} showed that a simple sum-pooling method working on deep convolutional features is superior to traditional aggregation methods working on shallow features \cite{ref28}; Su \textit{et al.} proposed the multi-view CNN (MVCNN) model, which adopts max pooling to combine information of photos captured from multiple orientations for 3D recognition \cite{ref23}; Wang \textit{et al.} introduced three popular pooling functions in their proposed mi-net, i.e. max pooling, mean pooling, and log-sum-exp (LSE) pooling (a smooth version and convex approximation of the max pooling) \cite{wang2018revisiting}; and Kraus \textit{et al.} proposed the Noisy-AND pooling function to handle outliers \cite{10.1093/bioinformatics/btw252} .

A major flaw of pooling is its incapability to focus on important instances, while the attention mechanism is a good choice, as it can assign scores/weights to instances. Till now, various attention-based feature aggregation methods have been developed. Yang \emph{et al}. proposed a neural aggregation network for video face recognition, NAN\cite{ref24}, which aggregates multiple features by an attention module. Ilse \textit{et al.} proposed a general attention-based deep MIL framework, which introduced both attention and gated-attention functions \cite{ilse2018attention}. Due to their good interpretability, attention-based methods have gained more popularity in biomedical image processing, \textit{e.g.} Yao \textit{et al.} used the attention-based MIL pooling for whole slide image analysis. In addition, the attention-based MIL can be implemented in various types of neural networks. For instance, Annofly uses long short-term memory model as the backbone network \cite{annofly}, while Set Transformer \cite{pmlr-v97-lee19d} and ImPLoc \cite{long2020imploc} are based on Transformer models. Besides these two mechanisms, there are some other deep MIL models with special designs, like DeepMIML \cite{feng2017deep} utilizing a pluggable sub-concept layer to learn the latent relationship between input patterns and output semantic labels. And Tu \textit{et al.} proposed a graph neural network-based model, which regards each instance as a node in the graph. Learning the graph embedding is essentially an information aggregation process \cite{tu2019multiple}.

The proposed HAMIL model has a different aggregation mechanism compared with the existing methods.
It has trainable and non-linear convolution aggregation units, and designs a hierarchical clustering protocol for permutation-invariance. Different from the hierarchical graph clustering in GNN-based model \cite{tu2019multiple}, both the numbers of clusters and aggregation times are automatically determined by the size of input bag rather than fixed hyperparameters.
\vspace{-1mm}
\section{Methods}

\subsection{Problem description}
In this study, we discuss a more complex MIL problem, namely multi-instance  multi-label learning. Formally, let $\mathcal{X}$ denote the sample set, i.e. $\mathcal{X}=\{X_i\}$, where $i \in \{1,2,...,n\}$, $n$ is the number of samples, and $X_i$ is a sample; $X_i=\{x_{i,1}, x_{i,2}, \ldots, x_{i,m}\}$, where $m$ is the number of instances of the $i$th sample, $x_{i,j}$ ($j \in \{1, 2, ..., m\}$) denotes an instance of $X_i$. Let $\mathcal{Y}=\{Y_i\}$ be the output space, and $Y_i=\{y_1,y_2, \ldots, y_k\}$ corresponds to the label set of $X_i$. The goal is to learn a mapping function $f:\mathcal{X}\mapsto \mathcal{Y}$. Especially, as we focus on the image processing tasks, here each instance is an image and each sample is represented by a bag of images. 

\subsection{Model architecture}
To process input bags of any size and varying sizes, HAMIL sets no limit on the number of instances for a bag and implements a hierarchical aggregation protocol. 
The model architecture is shown in Figure \ref{fig:framework}. There are three main components, namely feature extraction, feature aggregation, and classification. The first component consists of several convolutional layers serving as the feature extractor. The CNN layers are followed by a hierarchical aggregation procedure to produce a unified representation for the input set of instances, which is further fed to the fully connected layer for classification.

\renewcommand{\algorithmicrequire}{ \textbf{Input:}}      
\renewcommand{\algorithmicensure}{ \textbf{Output:}}     
\begin{algorithm}
  \caption{Construction of the hierarchy of instances} \label{algo1}
  \label{alg:Framwork}
  \begin{algorithmic}[1]
     \REQUIRE
A bag of images, $X$. \\
$X=\{x_{1}, x_{2}, \ldots, x_{m}\}$, where $m$ is the number of instances of $X$.
\ENSURE
A queue recording the aggregation history.
\STATE Let $\mathcal{S}$ be a set of clusters. 
\STATE $\mathcal{S}=\{\{x_1\}, \{x_2\}, \cdots, \{x_m\}\}$.
\STATE $I_{\{x_i\}} = i$, $maxI$ = $m$, where $I$ denotes the index of clusters and $maxI$ is the maximum index of clusters in $\mathcal{S}$.
\STATE Let $\mathcal{T}$ be a queue of triplets as defined in Eq. (\ref{eq:triplet}). Initialize $\mathcal{T}$ to an empty queue.
\WHILE{$|S|>1$}%
    \STATE $n=|S|$.
    \STATE Let $minL$ be the minimal pairwise distance between clusters. $minL = +\infty$.
    \STATE Let ${C1}, {C2}$ be the two clusters to be merged.
    \FOR{$i=1$ to $n-1$}
        \FOR{$j=i+1$ to $n$}
  \STATE $L_{i,j} = dis(S_i,S_j)$,where $dis$ is a distance function and $S_i$ is the $i$ th element in $\mathcal{S}$ by the ascending order of indexes.\label{line}
            \IF{$L_{i,j} < minL$}
                \STATE $minL = L_{i,j}$.
                \STATE ${C1}$ = $S_i$,  ${C2}$ = $S_j$.
            \ENDIF
        \ENDFOR
    \ENDFOR
    \STATE $C$ = ${C1} \cup {C2}$
    \STATE $S = (S\cup C)\setminus {C1}\setminus {C2} $.
    \STATE  $maxI = maxI+1$, $I_{C} = maxI$.
    \STATE Define a triplet $t = <I_{{C1}}, I_{{C2}}, I_{{C}}>$. 
\STATE Push $t$ into $\mathcal{T}$.   %
\ENDWHILE
\STATE Return $\mathcal{T}$.
\end{algorithmic}
\end{algorithm}

Specifically, by regarding the input bag as a set of images, HAMIL constructs a hierarchy of images/instances within each bag, thus determines an aggregation order according to the tree structure, i.e. from the leaf node to the root node of the hierarchical tree. The construction of instance hierarchy is described in Algorithm \ref{algo1}. In Line \ref{line}, the distance between two clusters is defined as the minimum distance of all cross-cluster instance pairs, i.e. the single-link method. And the distance between instances is Euclidean distance computed based on features yielded by the convolutional layers in the first component of HAMIL.

The hierarchy of instances is a binary tree in nature. During the construction of the hierarchy, we record the merging history of clusters. Assume that there are a total of $K$ merging steps (note that $K$ is not a hyperparameter but the height of the binary tree determined by the number of instances), then we will keep a queue $\mathcal{T}$ of length $K$. Each element in the array is a triplet denoted by $t_k$, i.e.,
\begin{equation}
    t_k = <I_{{C1}}, I_{{C2}}, I_{{C}}>,\\
k \in \{1, \cdots, K\}
\label{eq:triplet}
\end{equation}
$t_k$ consists of three indexes of clusters in $S$. The first two are the indexes of the two clusters being merged in the $k$-th step, and the last one is the index of the newly generated cluster. 
Then, the aggregation order is determined by the records in $\mathcal{T}$, and the specific aggregation operations are defined by kernel functions as described in the next Section.

\subsection{Feature aggregation unit}\label{sec:fa}

\begin{figure}
\begin{center}
  \includegraphics[width=\linewidth]{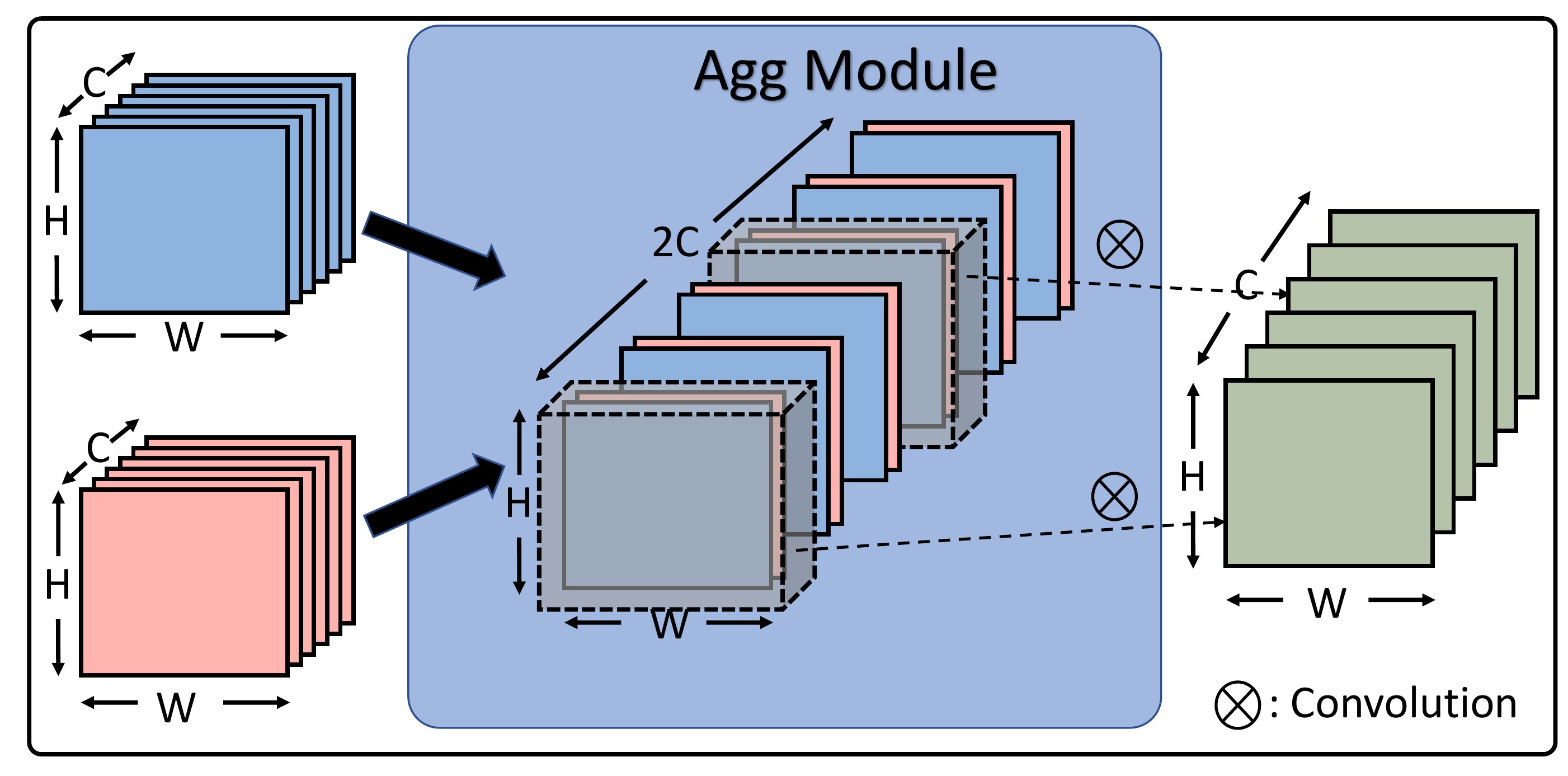}
\end{center}
   \caption{Illustration of feature aggregation in HAMIL.
   }
\label{fig:units}
\end{figure}

As the convolution operation can be regarded as a weighted average of inputs, we design the aggregation units via convolution. In detail, given two input feature maps of size $H\times W\times C$, which can be considered as $C$ pairs of $H\times W$ matrices along the channels, i.e. for each channel there is a pair of matrices. We first input each pair into the feature aggregation unit to obtain the aggregated output of size $H\times W$. Then, the $C$ outputs are concatenated into the final output of size $H \times W \times C$. Figure \ref{fig:units} shows an illustration of the aggregation units. 

Formally, let $x_{1}$ and $x_{2}$ be the feature maps to be aggregated. The aggregation is defined in Eq. (\ref{eq:agg}),
\begin{equation}
\begin{split}
    &\textbf{X}=[x_{1}, x_{2}],\\ 
    &O= \textbf{W}*\textbf{X}+b,
\end{split}
\label{eq:agg}
\end{equation}
where $\textbf{X}$ is a tensor composed by the feature maps, $*$ is the convolution operator, $\textbf{W}$ is the convolutional filter, $b$ is the bias, and $O$ is the aggregated feature map. This is a one-layer aggregation unit, We call it L1Agg. 
To meet the needs of more complex tasks, the aggregation units can be extended to deeper versions with multiple layers. Eqs. (\ref{eq:l2}) and (\ref{eq:l3}) formulate the two-layer and three-layer aggregation units which nest the basic aggregation operation twice and three times, respectively.
 
\begin{equation}
O= \textbf{W}'*f(g(\textbf{W}*\textbf{X}+b))+b',
\label{eq:l2}
\end{equation}
where $g(\cdot)$ is a normalization function, $f(\cdot)$ denotes the ReLU function.
\begin{equation}
O= \textbf{W}''*f(g(\textbf{W}'*f(g(\textbf{W}*\textbf{X}+b))+b'))+b'',
\label{eq:l3}
\end{equation}

No matter how many layers are there, the aggregation units are shared by all aggregation operations. Thus, the aggregation module introduces only a slight increase in the number of parameters compared to the backbone CNN model.

\section{Experiments} 
We assess the model performance on two large-scale image classification tasks, involving two typical kinds of bio-images, i.e. microscopic cell images and gene expression images. The two tasks are described below.
\vspace{1mm}
\newline 
\textbf{Task I: Prediction of protein subcellular location using immunofluorescence (IF) images. }Each protein corresponds to a bag of microscopy images captured from multiple tissues. The labels, i.e. cellular locations, are predicted based on all localization patterns implied in these images. A protein may exist in multiple locations.
\vspace{1mm}
\newline
\textbf{Task II: Gene function annotation using gene expression images}. The \textit{in situ} hybridization (ISH) imaging technology visualizes spatial distribution patterns of gene expression in tissues and help to reveal gene functions. Each gene corresponds to a bag of expression images captured in different angles or experimental trials. The labels are functional annotation terms. A gene may have more than one annotation terms.

Apparently, both of these two tasks are multi-instance multi-label classification. We compare HAMIL with both single-instance learning models and the existing feature aggregation models as listed in the following.
To assess the performance of HAMIL, we compare it with 7 baseline models, including:
\begin{itemize}
    \item A single-instance learning model (SI)\footnote{It has the same backbone network as HAMIL but its inputs are single images, which have the same labels as the bags they belong to. The prediction results of single images are combined per bag to yield the bag labels.};
    \item Three pooling-based methods, MI with mean pooling, MVCNN \cite{ref23} with max pooling, and SPoC \cite{ref28} with sum pooling;
    \item Two attention-based methods, NAN \cite{ref24} and Attention \cite{ilse2018attention};
    \item A specially designed deep MIL model, DeepMIML \cite{feng2017deep}. 
\end{itemize}



All the baseline models have the same backbone network (ResNet18) for extracting features from raw images. The prediction performance is evaluated by three metrics, AUC (the Area Under ROC Curve)\footnote{The averaged AUC value over labels, i.e. macro AUC}, macro F$_1$, and micro F$_1$.



\subsection{Task 1- Protein microscopy images}

\subsubsection{Data source} 
\begin{figure}[t]
\begin{center}
\includegraphics[width=0.9\linewidth]{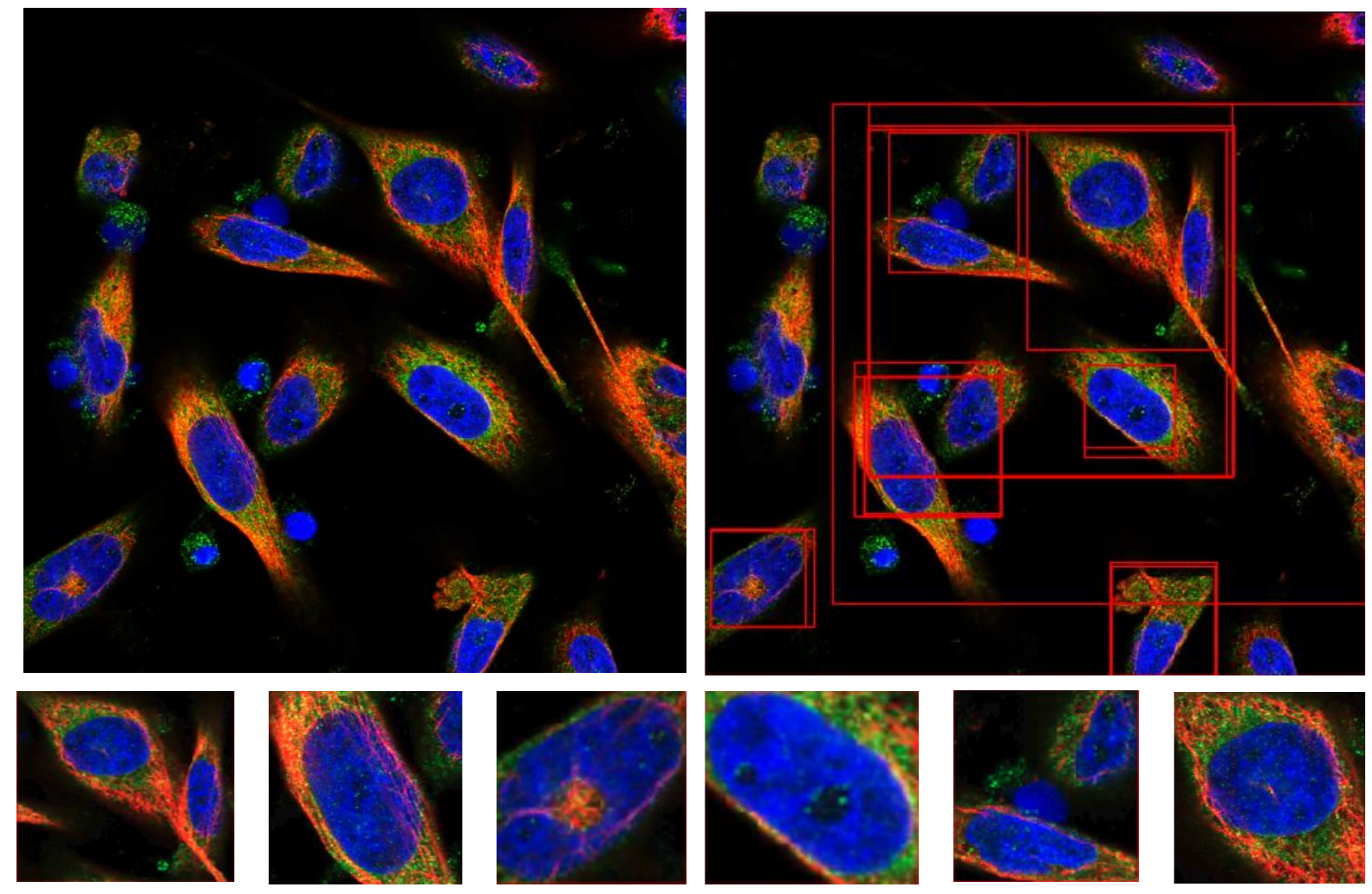}
\end{center}
\vspace{-1mm}
\caption{Extracted patches from HPA raw images. The top left figure is a raw image, the top right figure shows the results of selective search, and the bottom figures are the cropped and scaled patches. }
\label{fig:hpa}
\end{figure}

The data set was collected from the cell atlas of the human protein atlas (HPA) database \cite{uhlen2010towards}. We download 1600 bags of IF images, each of which corresponds to a protein. The location annotations of proteins are hand-crafted in the database. The total number of images is 19,777. The sizes of raw images include $800\times 800$, $1728\times 1728$, and $2048\times 2048$. Obviously, the large size of images and small numbers of bags bring great computation difficulties. Considering that there are usually multiple cells within an image, the selective search algorithm \cite{uijlings2013selective} is adopted to pick the cellular patches from raw images as shown in Figure \ref{fig:hpa}. We scale the selected patches into a fixed size $512\times 512$. Proteins with too many patches are split into multiple bags. The top-10 frequent classes are selected for our experiment. The training and test sets are partitioned at the protein level. Details about the data set can be found in suppl. Table \ref{tab:S1}.

\vspace{-2mm}

\subsubsection{Experiment details}Adam optimizer \cite{ref25} is adopted to optimize the network. The learning rate is set to 0.0001, and the parameters $\beta_1$ and $\beta_2$ of the Adam optimizer are set to 0.9 and the default value, respectively. The batch size is 8 and the loss function is standard BCE loss. We select the L1Agg function as our aggregation function, and the kernel size is set to 7, in addition, we use the BN layer after the convolution operation to normalize the feature map. The selected feature extraction network is ResNet18 \cite{he2016deep}, and the final output layer consists of 10 nodes (to predict 10 major cellular organelles).
\vspace{-2mm}
\subsubsection{Results}The prediction accuracy of HAMIL and 7 baseline models on Task I is shown in Table \ref{tab:base}.
\begin{table}
\small
\begin{center}
\caption{Performance comparison on Task 1$^*$}\label{tab:base}
\begin{tabular}{|l|l|r|r|r|}
\hline
\multicolumn{2}{|c|}{Method} &AUC &\small{MacroF$_1$} &\small{MicroF$_1$}\\
\hline\hline
\multicolumn{2}{|c|}{SI} & 0.938 &  0.572 & 0.734 \\
\hline
 \multirow{3}{*}{\small{Pool}}&MI & 0.927 & 0.577 & 0.745 \\
& MVCNN\cite{ref23} &0.938 & 0.675 & 0.763 \\
& SPoC\cite{ref28} & 0.941 & 0.714 & 0.770 \\
\hline
\multirow{3}{*}{\small{Attn}}&NAN\cite{ref24} & 0.928 & 0.705 & \textbf{0.784} \\
&\footnotesize{Attention}\cite{ilse2018attention}&  0.910	&0.559	&0.729\\
&\footnotesize{Gated-Attention}\cite{ilse2018attention}& 0.900	& 0.522	&0.724\\
\hline

\multicolumn{2}{|c|}{DeepMIML\cite{feng2017deep}}&0.915	&0.662&	0.746\\
\multicolumn{2}{|c|}{HAMIL (ours)} & \textbf{0.944}& \textbf{0.733}  & \textbf{0.784}\\
\hline
\end{tabular}
\end{center}
\begin{flushleft}\scriptsize
$^*$`Pool' denotes pooling-based aggregation and `Attn' denotes attention-based aggregation. Gated-Attention used gated-attention mechanism for aggregation. The result of DeepMIML is obtained by resizing the images to 224$\times$224, which leads to much higher accuracy than using the original size. All methods use ResNet18 as feature extractor.
\end{flushleft}
\end{table}
As can be seen, HAMIL achieves the best performance. SI treats the task as single-instance learning, which assumes that every single instance has the same label set as the whole bag, thus would introduce a lot of noisy samples. 

The three pooling methods, MI, MVCNN and SPoC, adopt mean, max, and sum pooling, respectively. SPoC achieves the best performance, perhaps due to the `centering prior' incorporated in the sum pooling, which assigns larger weights to center area of feature maps. As preprocessed by selective search (Fig. \ref{fig:hpa}), the central regions of most images contain more cell information.

As for the attention-based methods, NAN performs much better than `attention' and `gated-attention'. Despite different model settings and equations for computing attention scores, the major difference is that we apply NAN on feature maps while the latter two methods on feature vectors. Following the implementation in \cite{ilse2018attention}, the feature maps yielded by CNN layers are turned into lower-dimensional feature vectors via fully connected layers, thus the attention and gated-attention mechanisms work on feature vectors. This FC transformation may result in information loss.

DeepMIML does not obtain a comparable result to HAMIL, perhaps because it was designed to identify multiple concepts in single-image input, which is another kind of MIL problem different from ours.



\subsection{Task 2 - Gene expression images}
\subsubsection{Data source} 
\begin{figure}[t]
\begin{center}
\includegraphics[width=0.9\linewidth]{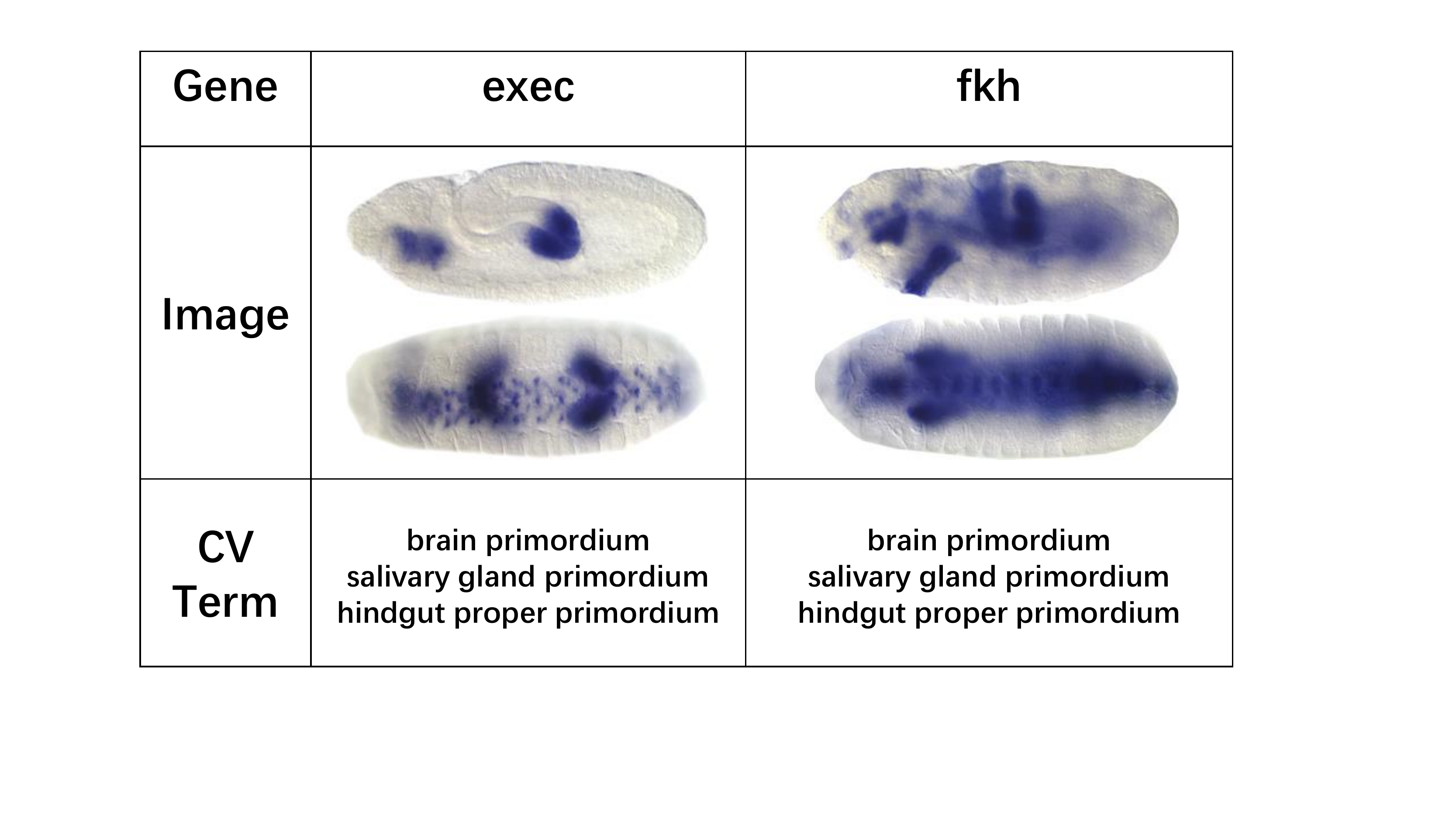}
\end{center}
\vspace{-1mm}
\caption{Examples of the gene annotation data of \emph{Drosophila} embryos. The first row shows gene names, the second row shows gene expression images for each gene, and the last row shows the corresponding CV terms of the gene, i.e. labels.}
\label{fig:fly}
\end{figure}

The data set consists of standard gene expression images of \emph{Drosophila} embryos from the FlyExpress database~\cite{kumar2011flyexpress}. All the images were extracted from the Berkeley \textit{Drosophila} Genome Project (BDGP) (\url{www.fruitfly.org})~\cite{tomancak2002systematic,tomancak2007global} and preprocessed to a uniform size $180\times 320$. The data set is in bag-level. Each bag of images presents the expression distribution on the \emph{Drosophila} embryo for a single gene, in which the images were captured from different orientations, i.e. dorsal, lateral and ventral. We use the same dataset as FlyIT \cite{ref27}, where the dataset is divided (at bag-level) into training, validation, and test sets according to the ratio of 4:1:5 (suppl. Table \ref{tab:S1} shows data statistics). The total number of labels is 10, each of which is an ontology term describing anatomical and developmental properties. This task is also a multi-label classification problem, as each gene may have multiple developmental terms. Figure \ref{fig:fly} shows some example images.

\vspace{-2mm}
\subsubsection{Experiment details}The experimental settings in Task 2 are almost the same as in Task 1, except that the batch size is set to 32, because the gene expression images are smaller than protein microscopy images, so we can add more train samples during one train process. Here we compare HAMIL with additional four state-of-the-art MIL methods proposed for the \textit{Drosophila} embryo image analysis \cite{li2012drosophila,annofly,ref27}.
E-MIMLSVM is an MIML algorithm that adapts the kernel function of support vector machines to the MIML scenario~\cite{li2012drosophila}. 
FlyIT\cite{ref27} adopts image stitching to combine raw images. AnnoFly \cite{annofly} first extracts features via ResNet, then employs RNN to deal with the multi-instance input. 
\vspace{-2mm}
\subsubsection{Results}The experimental results are shown in Table \ref{tab:task2}. In this task, the image quality is much higher than that of Task I and the number of instances per bag is fewer. That could explain why pooling-based methods achieve better results on this Task.
By contrast, Task I has much more instances per bag and highly varying image quality (see examples in Fig. \ref{fig:vis2}), the attention mechanism which identifies high-quality instances can help make better decision based on the important instances.

Interestingly, the pooling and attention-based methods have comparable and even better performance than the previously proposed MIL methods for this task. E-MIMLSVM is a shallow learning model. Annofly uses the pre-trained (on Imagenet) CNN to extract feature embeddings and then feeds them into an LSTM model. FlyIt conducts aggregation at the raw-image level. Neither of them fully exploit the feature learning ability of deep CNNs during the aggregation process.

In summary, the advantages of HAMIL over these state-of-the-art methods can be attributed to the end-to-end training of ResNet18 for feature extraction and the hierarchical aggregation operations on the feature maps.



\begin{table}
\footnotesize
\begin{center}
\caption{Performance comparison on Task 2$^*$}\label{tab:task2}
\begin{tabular}{|l|l|r|r|r|}
\hline
\multicolumn{2}{|c|}{Method} &AUC &\small{MacroF$_1$} &\small{MicroF$_1$}\\
\hline\hline

\multicolumn{2}{|c|}{SI} & 0.936 &  0.691 & 0.685 \\
\hline
\multirow{3}{*}{\small{Pool}}&MI & 0.938 & 0.688 & 0.706 \\
& MVCNN\cite{ref23} &0.942 & 0.741 & 0.743 \\
&SPoC\cite{ref28} & 0.944 & 0.729 & 0.737 \\
\hline
\multirow{3}{*}{\small{Attn}}&NAN\cite{ref24} & 0.927 & 0.726 & 0.734 \\
&\footnotesize{Attention}\cite{ilse2018attention}&  0.935	&0.692	&0.707\\
&\footnotesize{Gated-Attention}\cite{ilse2018attention}& 0.935	& 0.701	&0.701\\
\hline
\multicolumn{2}{|c|}{DeepMIML\cite{feng2017deep}}&0.922&	0.681&	0.700\\
\hline
\multicolumn{2}{|c|}{E-MIMLSVM\cite{li2012drosophila}} & 0.846 & 0.598 & 0.640 \\
\multicolumn{2}{|c|}{AnnoFly\cite{annofly}}&0.937&0.702&0.713\\
\multicolumn{2}{|c|}{FlyIT\cite{ref27}} & 0.936& 0.718 & 0.711 \\ 
\multicolumn{2}{|c|}{HAMIL (ours)} & \textbf{0.944} & \textbf{0.755} & \textbf{0.746} \\
\hline
\end{tabular}
\end{center}
\begin{flushleft}\scriptsize
$^*$The results of E-MIMLSVM$^+$, AnnoFly, and FlyIT are from \cite{ref27}.
\end{flushleft}
\end{table}
\vspace{-2mm}
\subsubsection{Visualization analysis} 
\begin{figure}[ht]
\centering
\subfigure[Raw image features]{\includegraphics[height=4cm,width=4cm]{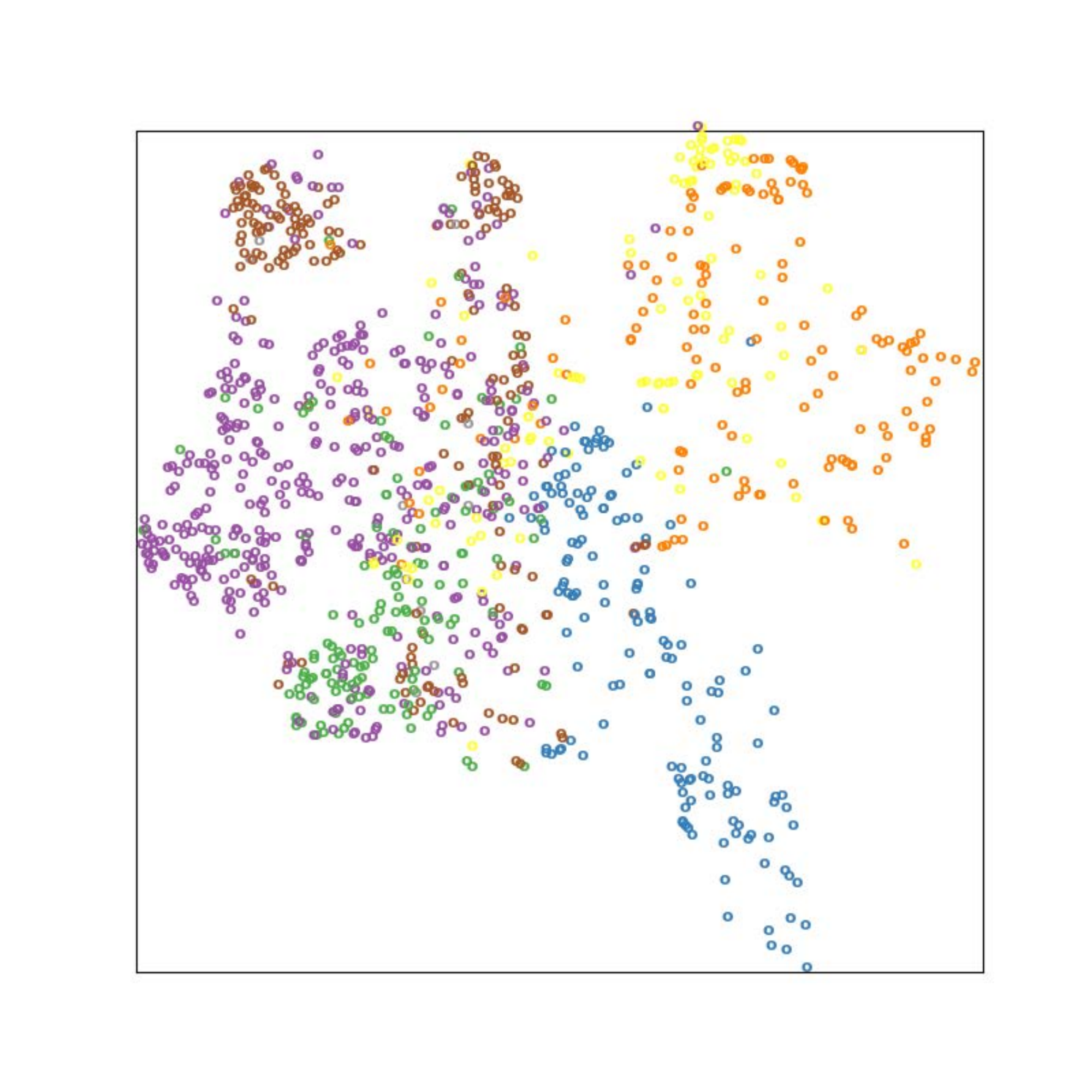}}
\subfigure[Aggregated features]{\includegraphics[height=4cm,width=4cm]{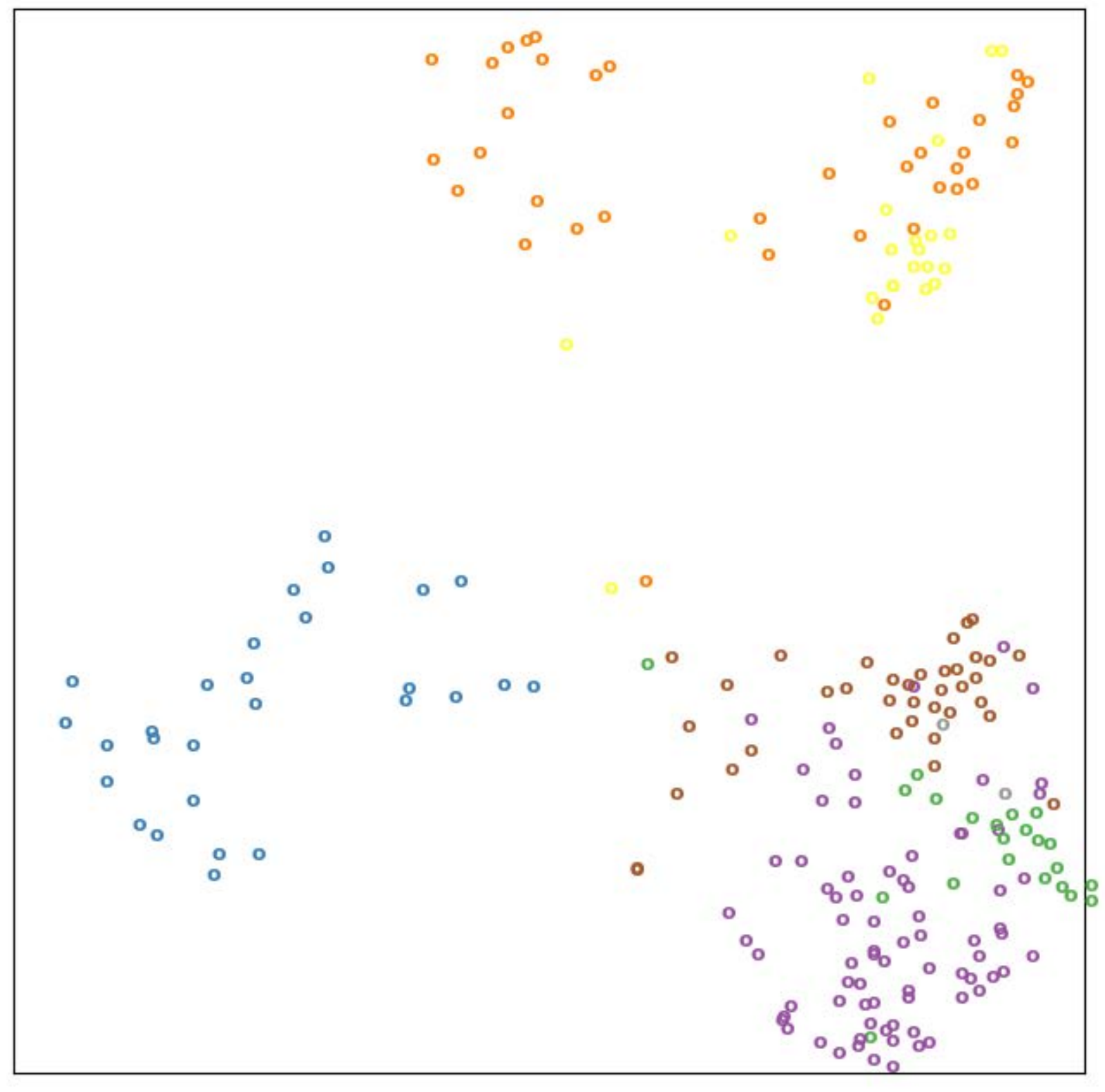}}
\caption{ Visualization of feature representations before and after aggregation. }
\label{fig:fig}
\end{figure}

We visualize the features representations with and without aggregation by projecting them onto a 2D space via tSNE algorithm \cite{tsne}. Figure \ref{fig:fig} (a) shows the 2D distribution of features learned by ResNet without aggregation, and Figure \ref{fig:fig} (b) shows the 2D distribution of aggregated features. Different colors denote different classes. The features of different categories before aggregation have a small margin and large overlap, which brings great difficulties to subsequent classifiers. After aggregation, samples of different categories are much more separated, with small overlap, which is beneficial for subsequent classification.



\subsection{Ablation Study and Analysis}
\subsubsection{Ablation of the hierarchical aggregation}
In order to verify the contribution of hierarchical design to the performance of HAMIL, we remove the hierarchical clustering in HAMIL and aggregate the instances in random order using the same convolutional aggregation functions, which is denoted by RAMIL. As RAMIL is not permutation-invariant, during its training process, we randomly shuffle the order of instances within bags in each epoch. The comparison results of HAMIL and RAMIL are shown in Table \ref{tab:ablation1}. HAMIL outperforms RAMIL on all the three metrics, suggesting that the hierarchical aggregation scheme is able to improve the performance of the model.

\begin{table}
\footnotesize
\begin{center}
\caption{Performance comparison on Task 2$^*$}\label{tab:ablation1}
\begin{tabular}{|c|c|r|r|r|r|}
\hline
Task&Method & AUC& MacroF$_1$& MicroF$_1$\\
\hline\hline
\multirow{3}{*}{Task I}& RAMIL&0.935	&0.720&	0.778\\
    &HAMIL-A & 0.939 & 0.695 & 0.773 \\
      &HAMIL&\textbf{0.944}&\textbf{0.733}&\textbf{0.784}\\
      \hline
\multirow{3}{*}{Task II}&RAMIL& 0.935 & 0.735 & 0.730 \\
&HAMIL-A & 0.932 & 0.736 & 0.732 \\
&HAMIL & \textbf{0.944} &\textbf{0.755} & \textbf{0.746} \\
\hline
\end{tabular}
\end{center}
\begin{flushleft}\scriptsize
$^*$RAMIL and HAMIL are two variants of HAMIL. RAMIL performs aggregation on random-ordered instances but with the same aggregation units as HAMIL; HAMIL-A adopts the same hierarchical aggregation protocol but replaces the convolutional aggregation units with mean pooling.
\end{flushleft}
\end{table}

\begin{figure*}
\begin{center}
\includegraphics[width=\linewidth]{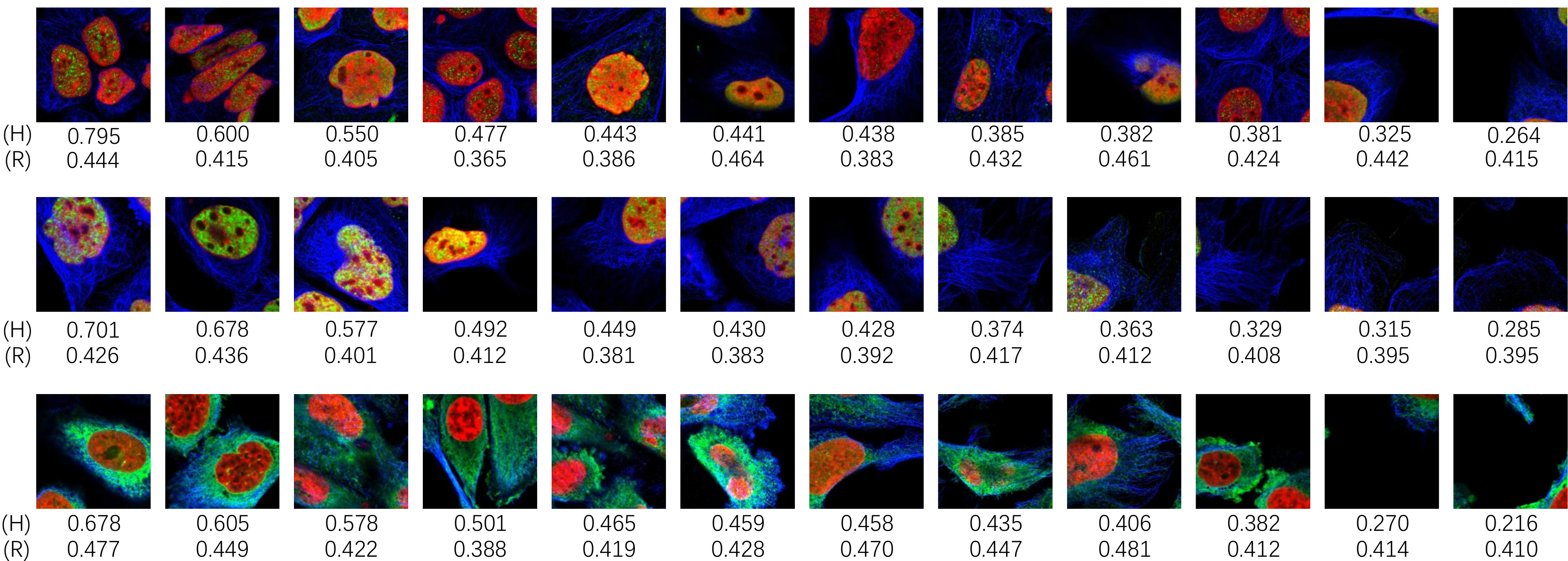}
\end{center}
\vspace{-2mm}
\caption{
Visualization of scores for image instances. The three rows denote three bags of images, and the numeric values are the scores assigned by HAMIL and RAMIL to the corresponding images.
}
\label{fig:vis2}
\end{figure*}

In addition, to get more insight on the aggregation operation, we compute the cosine similarity between aggregated features and the features of input images, and take the similarity value as the score of the input images, which can be regarded as a kind of preference that model assigns to the instances. Figure \ref{fig:vis2} visualizes the images with different scores. Each row shows images from the same bag, the two numbers below each image represent the weights assigned to the image by HAMIL and RAMIL, respectively. As can be seen, the images with high scores contain relatively obvious local patterns corresponding to the target labels, suggesting that the feature aggregation retains more information from the high-quality images, i.e.  the features aggregated by HAMIL will automatically focus on important features. 
By contrast, it can be found that RAMIL pays nearly equal attention to different quality images, as the scores differ slightly. This is because RAMIL aggregates each input in a random order, i.e., each time it randomly selects two feature embeddings to aggregate, thus there is no mechanism for attention allocation. As a result, when the quality of images varies a lot, the performance of RAMIL cannot be guaranteed.

\subsubsection{Ablation of the convolutional aggregation units}\label{sec:abl2}
To verify the effectiveness of convolutional aggregation units in our design, we use a simple average operation to replace the non-linear convolutional aggregation unit. HAMIL-A is used to denote this alteration of aggregation unit. In Table \ref{tab:ablation1} we compare it with HAMIL in Task I and II. Because convolutional aggregation unit is a trainable module, its performance is better than simple average operation. Suppl. Table \ref{tab:ablation2} also shows the results on localized content-based image retrieval.

\section{Comparison on aggregation mechanisms}
HAMIL is a new aggregation framework designed for addressing MIL in image processing. 
It allows variable sizes of input bags without a restriction on the bag size, and it has a light-weight architecture with simple aggregation operations. The hierarchical aggregation protocol makes the model invariant to the order of instances in bags.  
Actually, many aggregation methods also have these advantages, like pooling and attention-based \cite{wang2018revisiting,ilse2018attention}. 
A comparative analysis of these three mechanisms is listed below.

i) Traditional pooling-based methods (max, sum or mean) are non-trainable, thus have limited learning ability; while both attention-based aggregation function and HAMIL are trainable.

ii) The existing attention-based aggregation methods assign weight scores to instances, i.e. they treat an image/instance as a whole, no matter how the pairwise instance similarity is computed (based on feature maps or feature embedding vectors). By contrast, both pooling-based and HAMIL can operate on local regions of multiple instances. 
Although attention methods have good interpretability (the weights directly reflect the importance of instances), the weighted-sum aggregation function is incompetent to express complex aggregation mechanisms. 

iii) According to the experimental results, the pooling-based methods are very efficient and effective on data sets with high quality, and attention-based methods have more advantages in handling data with much noise or varying quality; while HAMIL is applicable to both cases, due to the powerful learning ability of its aggregation units.   



Although HAMIL is designed for handling raw image data, we perform an additional experiment to investigate its performance on traditional MIL data sets (as shown in suppl. Table \ref{tab:mil-datasets}). As the input features are provided (preprocessed via feature engineering), we remove the first component of HAMIL, i.e. the CNN layers for learning image features. Following the practice in \cite{ilse2018attention}, We run the experiments using 10-fold cross-validations for 5 times (each time the data set is randomly partitioned into 10 folds). Experimental details can be found in suppl. Section \ref{section:details-mil}.

Suppl. Table \ref{tab:ablation2} shows the results on three localized content-based image retrieval tasks, and suppl. Table \ref{tab:ablation-drug} shows the results for two drug activity data sets. Considering that the input consists of statistical features of images, we not only replace the original 2D conv to 1D conv but also add HAMIL-A (as mentioned in Section \ref{sec:abl2}) into comparison. As can be seen, HAMIL achieves a little advantage on Fox and Elephant data sets while no advantage on others. Besides, HAMIL-A and HAMIL have very close performance.

From these experimental results, it can be seen that as the aggregation units of HAMIL are convolution-based and they directly function on feature maps, HAMIL is more suitable for processing raw image data. Nevertheless, the proposed HAMIL model could be an alternative feature aggregation protocol applied to the data sets with traditional features, which obtains comparable results with the existing aggregation methods. The contributions of HAMIL can be summarized as follows.

i) The convolution-based aggregation function of HAMIL is able to capture local neighborhood information in images during the aggregation process and learn complex non-linear aggregation patterns.
  
ii) The hierarchical design makes the model insensitive to the order of instances and improves model accuracy.
    
iii) The end-to-end architecture further improves the feature learning, as the hierarchical clustering can be dynamically adjusted to yield better clustering of instances based on the updated feature representations.

\section{Conclusion}
We propose a hierarchical aggregation network for multi-instance learning in image classification. Different from the mainstream pooling and attention-based methods, the proposed HAMIL utilizes convolution operations for aggregation on feature maps and a hierarchical architecture to ensure permutation-invariance. HAMIL achieves better performance than the existing aggregation methods on two microscopy image classification tasks. Moreover, as a general feature aggregation network, HAMIL can be easily applied to other MIL image processing tasks.


%

\appendices

\setcounter{equation}{0}
\setcounter{figure}{0}
\setcounter{table}{0}
\setcounter{section}{0}
\makeatletter
\renewcommand{\theequation}{S\arabic{equation}}
\renewcommand{\thefigure}{S\arabic{figure}}
\renewcommand{\thetable}{S\arabic{table}}
\renewcommand\thesection{\Alph{section}}

\section{Details for Tasks I and II}\label{section:details-hpa-fly}
\renewcommand{\thetable}{S\arabic{table}}

\begin{table}[H]
\footnotesize
    \centering
   \caption{Data Overview of Tasks I and II}\label{tab:S1}
 \begin{tabular}{|l|r|r|r|r|r|r|}
    \hline
\multirow{2}{*}{Dataset}& \multicolumn{3}{c|}{Protein microscopy image}& \multicolumn{3}{c|}{Gene expression image}\\
\cline{2-7}
        &Train&Val&Test&Train&Val&Test\\
        \hline\hline
Bag \#&5496&1320&6901&2714&678&3393\\
Instance \#&69943&16687&86964&10237&2426&12872\\
\hline
\end{tabular}
\end{table}
\vspace{-2mm}

The datasets can be accessed at \url{https://bioimagestore.blob.core.windows.net/dataset?restype=container&comp=list}

\section{Details for classical MIL data sets}\label{section:details-mil}


\begin{table}[H]
    \centering
   \caption{Overview of five classical MIL data sets}\label{tab:mil-datasets}
 \begin{tabular}{|l|r|r|r|}
\hline
  Dataset &\# of bags &\# of instances &\# of featrues \\
  \hline\hline
  Musk1 & 92 & 476 & 166 \\
  Musk2 & 102 & 6598 & 166 \\
  Tiger & 200 & 1220 & 230 \\
  Fox & 200 & 1302 & 230 \\
  Elephant & 200 & 1391 & 230 \\
\hline
\end{tabular}
\end{table}

\begin{table}[H]
    \centering
   \caption{Model architecture of HAMIL and HAMIL-A working on the five traditional data sets$^*$}\label{tab:model-arch}
 \begin{tabular}{|l|c|}
\hline
  Layer & Type  \\
  \hline\hline
  1 & fc-256+ReLU \\
  2 & dropout \\
  3 & fc-128+ReLU \\
  4 & dropout \\
  5 & fc-64+ReLU \\
  6 & dropout \\
  7 & hierarchical clustering \\
  8 & conv/average aggregation \\
  9 & fc-1+sigm \\
\hline
\end{tabular}
\begin{flushleft}\scriptsize
$^*$To compare with previous studies, Layers 1-6 and 9 have exactly the same settings as Ilse et al's attention-based method \cite{ilse2018attention}.\\
\end{flushleft}
\end{table}

\begin{table}[H]
\footnotesize
\begin{center}
\caption{Performance comparison of classification accuracy (mean$\pm$std) on localized content-based image retrieval$^1$}\label{tab:ablation2}
\begin{tabular}{|l|l|l|l|}
\hline
Method & Fox & Tiger & Elephant\\
\hline\hline
mi-SVM\cite{2002mi-svm}  & 0.582 & 0.784 & 0.822  \\
MI-SVM\cite{2002mi-svm}  & 0.578 & 0.840 & 0.843 \\
MI-Kernel\cite{2002mi-kernel}  & 0.603 & 0.842 & 0.843 \\
EM-DD\cite{2001em-dd}  & 0.609$\pm$0.101 & 0.730$\pm$0.096 & 0.771$\pm$0.097\\
mi-Graph\cite{zhou2009multiinstance}  & 0.620$\pm$0.098 & \textbf{0.860$\pm$0.083} & 0.869$\pm$0.078\\
miVLAD\cite{2017Scalable}  & 0.620$\pm$0.098 & 0.811$\pm$0.087 & 0.850$\pm$0.080\\
miFV\cite{2017Scalable}  & 0.621$\pm$0.109 & 0.813$\pm$0.083 & 0.852$\pm$0.081\\
\hline\hline
mi-Net\cite{wang2018revisiting}  & 0.613$\pm$0.078 & 0.824$\pm$0.076 & 0.858$\pm$0.083\\
MI-Net\cite{wang2018revisiting}  & 0.622$\pm$0.084 & 0.830 $\pm$0.072 & 0.862$\pm$0.077\\
MI-Net with DS\cite{wang2018revisiting}  & 0.630$\pm$0.080 & 0.845 $\pm$0.087 & 0.872$\pm$0.072\\
MI-Net with RC\cite{wang2018revisiting}  & 0.619$\pm$0.104 & 0.836 $\pm$0.083 & 0.857$\pm$0.089\\
\hline\hline
Attention\cite{ilse2018attention} & 0.615$\pm$0.043 & 0.839 $\pm$0.022 & 0.868$\pm$0.022\\
Gated-Attention\cite{ilse2018attention} & 0.603$\pm$0.029 & 0.845 $\pm$0.018 & 0.857$\pm$0.027\\
\hline\hline
HAMIL-A$^2$ & 0.631$\pm$0.117 & 0.806 $\pm$0.095 & \textbf{0.878$\pm$0.091}\\
HAMIL$^3$ & \textbf{0.647$\pm$0.105} & 0.815 $\pm$0.091 & 0.865$\pm$0.079\\
\hline
\end{tabular}

\end{center}
\begin{flushleft}\scriptsize
$^1$Results of the first 11 methods are from \cite{wang2018revisiting}; and results of the Attention and Gate-Attention methods are from \cite{ilse2018attention}.\\
$^2$ Replace the conv-based aggregation of HAMIL with mean pooling.\\
$^3$ Change the aggregation units of HAMIL from 2D conv to 1D conv.
\end{flushleft}
\end{table}

\begin{table}[H]
\footnotesize
\begin{center}
\caption{Performance comparison of classification accuracy (mean$\pm$std) on drug activity prediction$^*$}\label{tab:ablation-drug}
\begin{tabular}{|l|l|l|}
\hline
Method & MUSK1 & MUSK2 \\
\hline\hline
mi-SVM\cite{2002mi-svm}  & 0.874 & 0.836 \\
MI-SVM\cite{2002mi-svm}  & 0.779 & 0.843 \\
MI-Kernel\cite{2002mi-kernel}  & 0.880 & 0.893 \\
EM-DD\cite{2001em-dd}  & 0.849$\pm$0.098 & 0.869$\pm$0.108 \\
mi-Graph\cite{zhou2009multiinstance}  & 0.889$\pm$0.073 & \textbf{0.903$\pm$0.086}\\
miVLAD\cite{2017Scalable}  & 0.871$\pm$0.098 & 0.872$\pm$0.095\\
miFV\cite{2017Scalable}  & \textbf{0.909$\pm$0.089} & 0.884$\pm$0.094\\
\hline\hline
mi-Net\cite{wang2018revisiting}  & 0.889$\pm$0.088 & 0.858$\pm$0.110\\
MI-Net\cite{wang2018revisiting}  & 0.887$\pm$0.091 & 0.859$\pm$0.102\\
MI-Net with DS\cite{wang2018revisiting}  & 0.894$\pm$0.093 & 0.874$\pm$0.097\\
MI-Net with RC\cite{wang2018revisiting}  & 0.898$\pm$0.097 & 0.873$\pm$0.098\\
\hline\hline
Attention\cite{ilse2018attention} & 0.892$\pm$0.040 & 0.858$\pm$0.048\\
Gated-Attention\cite{ilse2018attention} & 0.900$\pm$0.050 & 0.863$\pm$0.042\\
\hline\hline
HAMIL-A & 0.892$\pm$0.120 & 0.857$\pm$0.102\\
HAMIL &  0.866$\pm$0.121 & 0.820$\pm$0.107 \\
\hline
\end{tabular}
\end{center}
\begin{flushleft}\scriptsize
$^*$Results of the first 11 methods are from Wang et al.(2018) \cite{wang2018revisiting}; and results of the Attention and Gate-Attention methods are from Ilse et al. (2018) \cite{ilse2018attention}.\\
\end{flushleft}
\end{table}
%

The experimental settings are almost the same for the five data sets.  We use the SGD optimizer, the momentum is 0.9, and the weight decay is 0.005. We set the learning rate to 0.0001. Our models are trained for 100 epochs on Fox, MUSK1, and MUSK2. As for Elephant and Tiger, the number of training epochs is 50, due to their faster convergence.



\ifCLASSOPTIONcaptionsoff
  \newpage
\fi


\bibliographystyle{IEEEtran}
\bibliography{ref}
\end{document}